\documentclass[twocolumn]{ceurart}
\usepackage{listings}
\usepackage[linesnumbered,ruled,vlined]{algorithm2e}
\SetKwInput{KwInput}{Input}
\SetKwInput{KwOutput}{Output}

\begin{document}


\copyrightyear{2022}
\copyrightclause{Copyright for this paper by its authors.
  Use permitted under Creative Commons License Attribution 4.0
  International (CC BY 4.0).}

\conference{CIKM-PAS'22: PRIVACY ALGORITHMS IN SYSTEMS (PAS) Workshop, Conference on Information and Knowledge Management,
  October 21, 2022, CIKM-PAS}

\title{$k$-Means SubClustering: A Differentially Private Algorithm with Improved Clustering Quality}


\author[1]{Devvrat Joshi}[%
email=devvrat.joshi@iitgn.ac.in,
]
\cormark[1]
\fnmark[1]
\address[1]{Indian Institute of Technology Gandhinagar, India}
\author[1]{Janvi Thakkar}[%
email=janvi.thakkar@iitgn.ac.in,
]
\cormark[1]
\fnmark[1]

\cortext[1]{Corresponding author.}
\fntext[1]{These authors contributed equally.}

\begin{abstract}
  In today's data-driven world, the sensitivity of information has been a significant concern. With this data and additional information on the person's background, one can easily infer an individual's private data. Many differentially private iterative algorithms have been proposed in interactive settings to protect an individual's privacy from these inference attacks. The existing approaches adapt the method to compute differentially private(DP) centroids by iterative Llyod's algorithm and perturbing the centroid with various DP mechanisms. These DP mechanisms do not guarantee convergence of differentially private iterative algorithms and degrade the quality of the cluster. Thus, in this work, we further extend the previous work on `Differentially Private $k$-Means Clustering With Convergence Guarantee' by taking it as our baseline. The novelty of our approach is to sub-cluster the clusters and then select the centroid which has a higher probability of moving in the direction of the future centroid. At every Lloyd's step, the centroids are injected with the noise using the exponential DP mechanism. The results of the experiments indicate that our approach outperforms the current state-of-the-art method, i.e., the baseline algorithm, in terms of clustering quality while maintaining the same differential privacy requirements. The clustering quality significantly improved by 4.13 and 2.83 times than baseline for the Wine and Breast$\_$Cancer dataset, respectively.
\end{abstract}


\maketitle

\section{Introduction}
Achieving extraordinary results is dependent on the data on which the machine learning models are trained. Data curators have a responsibility to provide datasets such that the privacy of data is not compromised. However, attackers use other public datasets to perform inference and adversarial attacks to get information about an individual in the dataset. Differential privacy is a potential technique for giving customers a mathematical guarantee of the privacy of their data\cite{dwork2008differential}. There are two fundamental settings in which differential privacy is used on data: in interactive setting data curator holds the data and returns the response based on the queries requested by third parties; while in non-interactive setting the curator sanitized the data before publishing\cite{narayanan2009data}.

Iterative clustering algorithms provide important insights about the dataset, which helps in a large number of applications. They are prone to privacy threats because they can reveal information about an individual with additional knowledge. Existing approaches obtain the set of centroids using Lloyd's K-means algorithm, then perturb them with a differentially private mechanism to add privacy \cite{lu2020differentially}. In contrast to Lloyd's K-means algorithm, which guarantees convergence, these algorithms do not provide any convergence guarantee. Getting differentially private centroids might not help in getting quality inferences because of this non-convergence. We studied an existing approach that provides this guarantee and converges in twice the number of iterations to Lloyd's algorithm while maintaining the same differential privacy requirements as existing works \cite{su2016differentially} \cite{lei2011differentially}. Their algorithm perturbs the centroids in a random direction from the center of the cluster. However, this lowers the quality of clustering, which is necessary for making inferences.

In this work, we propose a variant of the existing approach, which provides better clustering quality while using the same privacy budget. 
We used the intuition of Lloyd's algorithm that the next centroid will move in the direction where there is a higher number of data points.
Finally, we give the mathematical proof that our approach at any instance gives better clustering quality than the existing approaches. We have tested our approach on breat\_cancer, wine, iris, and digits datasets. We were able to get a significant improvement from the previous approach in terms of clustering quality.

Interactive setting implies that the dataset is not disclosed to the user, however, the data curator returns the response of each query received from the user by manipulating it using DP strategy.

Our main contribution includes: 
\begin{enumerate}
    \item We proposed SubClustering approach which has better clustering quality than the baseline (which is the current SOTA in terms of clustering quality). For the Wine and Breast$\_$cancer dataset, the clustering quality improved by 4.13 and 2.83 times respectively.
    \item In addition to improving the clustering quality, our algorithm used same privacy budget as that of the existing work. 
\end{enumerate}

\section{Related Work}
The concept of differential privacy has inspired a plethora of studies, particularly in the area of differentially private k-means clustering \cite{su2017differentially}\cite{dwork2011firm}\cite{mohan2012gupt} in an interactive setting. The important mechanisms of DP in the literature include: the Laplace mechanisms (LapDP) \cite{dwork2006calibrating}, the exponential mechanisms (ExpDP) \cite{mcsherry2007mechanism}, and the sample and aggregate framework \cite{nissim2007smooth}. To achieve differential privacy, many implementations included infusing Laplace noise into each iteration of Lloyd's algorithm. The proportion of noise added was based on a fixed privacy budget. Some of the strategies for allocating privacy budget included splitting the overall privacy budget uniformly to each iteration \cite{blum2005practical}. However, this requires us to calculate the number of iterations for the convergence, prior to the execution of algorithm, thus increasing the computational cost. Further, researchers overcome this weakness by allocating theoretically guaranteed optimal allocation method \cite{su2017differentially}, but the major assumption taken in this approach was that every cluster has the same size, which does not align with the real-world datasets. In another work, Mohan et al. \cite{mohan2012gupt} proposed GUPT, which uses Lloyd's algorithm for local clustering of each bucket where the items were uniformly sampled to different buckets. The final result was the mean of locally sampled points in each bucket with added Laplace noise. But, the clustering quality of GUPT was unsatisfying because a large amount of noise was added in the aggregation stage.

Based on the study of past literature on differentially private k-means clustering, Zhigang et al. \cite{lu2020differentially} concluded that convergence of an iterative algorithm is important to the clustering quality.
To solve this, they introduced the concept of the convergent zone and orientation controller. With the help of a convergent zone and orientation controller, they further create a sampling zone for selecting a potential centroid for the $i^{th}$ iteration. The approach iteratively adds noise with an exponential mechanism (ExpDP) by using prior and future knowledge of the potential centroid at every step of Lloyd's algorithm. The approach maintains the same DP requirements as existing literature, with guaranteed convergence and improvement in clustering quality. However, their algorithm perturbs the centroids in a random direction from the center of the cluster, degrading the quality of clustering. Thus, in this work, we further build upon the approach and significantly improve the clustering quality with the same epsilon privacy. 

\section{Preliminaries}
The definitions used in this work are briefly discussed in this section. The following is a formal definition of Differential Privacy:

\textbf{Definition 1 ($\epsilon$-DP \cite{dwork2006calibrating})}.
\textit{A randomised mechanism T is $\epsilon$-
differentially private if for all neighbouring datasets $X$ and $X'$ and for an arbitrary answer $s\in Range(T)$, T  satisfies 
$$ Pr[T(X)=s] \leq exp(\epsilon)\cdot Pr[T(X')=s],$$
where $\epsilon$ is the privacy budget.
}

Here, $X$ and $X'$ differ by only one item. Smaller values of $\epsilon$ imply a better privacy guarantee. It is because the difference between the two neighboring datasets is reflected by the privacy budget.
In this work, we use the ExpDP and LapDP. In exponential DP for non-numeric computation, they introduce the concept of scoring function $q(X,x)$, which represents the effectiveness of the pair $(X,x)$. Here $X$ is the dataset and $x$ is the response to the $q(X,x)$ on X.

The formal definition of Exponential DP mechanism is defined as follow:

\textbf{Definition 2 (Exponential Mechanism \cite{mcsherry2007mechanism})}.
\textit{Given a scoring function of a dataset $X, q(X,x),$ which reflects the quality of query respond x. The exponential mechanism T provides $\epsilon$-differential privacy, if $T(X) = \{ Pr[x] \propto exp(\frac{\epsilon \cdot q(X,x)}{2\Delta q}) \},$ where $\Delta q$ is the sensitivity of scoring function q(X,x), $\epsilon$ is the privacy budget.
}

\textbf{Definition 3 (Convergent \& Sampling Zones\cite{lu2020differentially})}. 
\textit{A region whose points satisfies the condition: \{ Node S: $\|S-{S_i}^{(t)}\| < \|{S_i}^{(t-1)}-{S_i}^{(t)}\|$\} is the convergent zone. ${S_i}^{(t)}$ is defined as the mean of ${C_i}^{(t)}$. A sub-region inside convergent zone is defined as a sampling zone.}

\textbf{Definition 4 (Orientation Controller\cite{lu2020differentially})}.
\textit{${X_i}^{(t)}$ is a direction from the center of the convergent zone to a point on its circumference. This is the direction along which the center of the sampling zone will be sampled, defined as the orientation controller.}

\section{Approach}
In this section, we explain our proposed approach and the baseline approach.

\subsection{Overview - KMeans Guarantee (Baseline)}
We took "Differentially Private K-Means Clustering with Convergence Guarantee" \cite{lu2020differentially} as our baseline and improved the clustering quality by further building on it. The key concept of the algorithm is to use ExpDP to introduce bounded noise into centroids at each iteration of Lloyd's algorithm. The technique is designed in a way that it ensures the new centroid is different from the centroid of Lloyd's algorithm while maintaining constraint given in \textbf{Lemma 1}. The constraint guarantees that the perturbed centroid will eventually converge with the centroid of Lloyd's algorithm.

\begin{figure}[t]
\includegraphics[width=\linewidth]{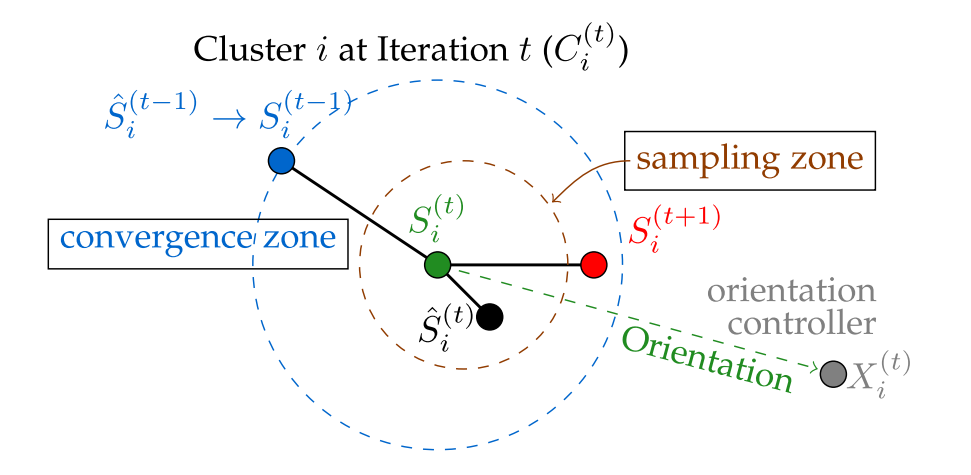}
\caption{Overview of KMeans Guarantee Approach}
\label{fig:kmeans.jpg}
\end{figure}

 Their algorithm has four main steps to update the centroids at each Lloyd step t \cite{lu2020differentially}. The overview of their approach can be seen in (\textbf{Figure : \ref{fig:kmeans.jpg}}).
\begin{enumerate}
    \item Let the differentially private centroid at iteration $t-1$ for a cluster $i$ be $\hat{S_i}^{(t-1)}$. Using this centroid, run one iteration of Lloyd's algorithm to get the current Lloyd's centroid ${S_i}^{(t)}$ for each cluster $i$.
    \item Using ${S_i}^{(t)}$ and ${S_i}^{(t-1)}$, generate a convergent zone for each cluster $i$ as described in \textbf{$Definition \:3$}.
    \item Generate a \textit{sampling zone} in the \textit{convergence zone}  and an \textit{orientation controller} ${X_i}^{(t)}$ for each cluster i as defined in \textbf{$Definition\: 3 \:and\:4$} respectively.
    \item Sample a differentially private $\hat{S_i}^{(t)}$ with ExpDP in the \textit{sampling zone} generated in step 3.
\end{enumerate}
The definition for the convergent zone (for convergence guarantee) and sampling zone (for centroid updating) is defined in Definition 3.
\begin{algorithm}[]
\DontPrintSemicolon
  
  \KwInput{$\textbf{X} = \{x_1,x_2, ...., x_N\}$: Dataset with N data points \\
  \textbf{k}: number of clusters \\
  $\epsilon^{exp}$: ExpDP privacy budget\\
  $\epsilon^{lap}$: Laplacian privacy budget for the converged centroids.\\
  $internalK$: number of sub-clusters per cluster\\
  }
  \KwOutput{\textbf{S:} Final clustering centroids}
Select $k$ centroids $\textbf{S}^{(0)} = ({S}^{(0)}_{1},{S}^{(0)}_{2}, ..., {S}^{(0)}_{k})$ uniformly from X.
\\
$iterationForLloyd$ = number of iterations to run the algorithm.
\\
 \For{iters i in $iterationForLloyd$}{
    \For{each Cluster i at Iteration t}{
        ${{C}^{(t)}_{i}} \leftarrow$ assign each $x_j$ to its closest centroid ${S_i}^{t-1}$; \\
        ${S_i}^{t} \leftarrow$ centroid of ${C_i}^{t}$;\\
        ${ConvergentZone_i}^{(t)} \leftarrow$ List of data points inside the spherical region having ${S_i}^{t}$ and ${S_i}^{t-1}$ as the endpoints of its radius.\\
        ${SamplingZone_i}^{(t)} \leftarrow$ run Algorithm 2 using ${ConvergentZone_i}^{(t)}$ , $internalK$;\\
        $\hat{S_i}^{(t)} \leftarrow$ sample from ${SamplingZone_i}^{(t)}$ using ExpDP with $q$ and $\epsilon^{exp}$;\\
        ${S_i}^{(t)} \leftarrow$ $\hat{S_i}^{(t)}$
    }
 }
 Publish: ${SamplingZone_i}^{(t)}$, $q$, $\epsilon^{exp}$, ${S_i}^{(t)}$\\
 $\textbf{S} \leftarrow$ add laplace noise with $\epsilon^{lap}$ to ${\textbf{S}}^{(t)}$;

\caption{Differentially Private $k-$Means SubClustering Algorithm}
\end{algorithm}

\begin{algorithm}[]
\DontPrintSemicolon
  
  \KwInput{\textbf{ConvergentZone}: Convergent Zone\\
  \textbf{internalK}: Subclustering K\\
  }
  \KwOutput{$SamplingZone^{t}_i$}
${\textbf{S}}^{(t)}$: Mean of ${ConvergentZone_i}^{(t)}$\\
$\mathbf{ConvergentZoneClusters} \leftarrow$ Cluster \textbf{ConvergentZone} using Lloyd's algorithm and $internalK$\\
$\mathbf{ConvergentZoneProbability} \leftarrow$ Assign probabilities to the $\mathbf{ConvergentZoneClusters}$ proportional to the number of points inside each cluster.\\
$\mathbf{{SamplingZone_i}^{(t)}} \leftarrow$ Sample a cluster from the $\mathbf{ConvergentZoneClusters}$ using $\mathbf{ConvergentZoneProbability}$\\

 Return: $\mathbf{{SamplingZone_i}^{(t)}}$;
\caption{SubClusterSamplingAlgorithm}
\end{algorithm}

\subsection{Overview - SubCluster Guarantee}

We build upon the KMeans Guarantee algorithm to achieve better clustering quality. Our idea differs from the baseline in terms of creating a sampling zone. For each cluster, we execute Lloyd's algorithm over its convergent zone to generate its sub-clustering. Further, we assign each sub-cluster with a probability linearly proportional to the number of points it contains. Finally, we sample the sub-cluster based on the assigned probability and define it as the sampling zone of the convergent zone. Drawing analogy from the KMeans Guarantee algorithm, our orientation controller is this sub-clustering and sampling technique. Intuitively, our algorithm ensures that the sampling zone lies towards the region containing a higher number of data points in an expected case. With this, we guarantee that our differentially private centroid moves in the direction where the number of data points is higher, incorporating the intuition of Lloyd's algorithm without compromising on the $\epsilon$-differential privacy. The probability of a differentially private centroid at $i-1^{th}$ iteration to move in the direction of a more populated region at the $i^{th}$ step of Lloyd's algorithm is also high. Thus, we introduce the concept of sub-clustering in the convergent zone and consequently sample one sub-cluster as our sampling zone.

\begin{figure}[t]
\includegraphics[width=\linewidth]{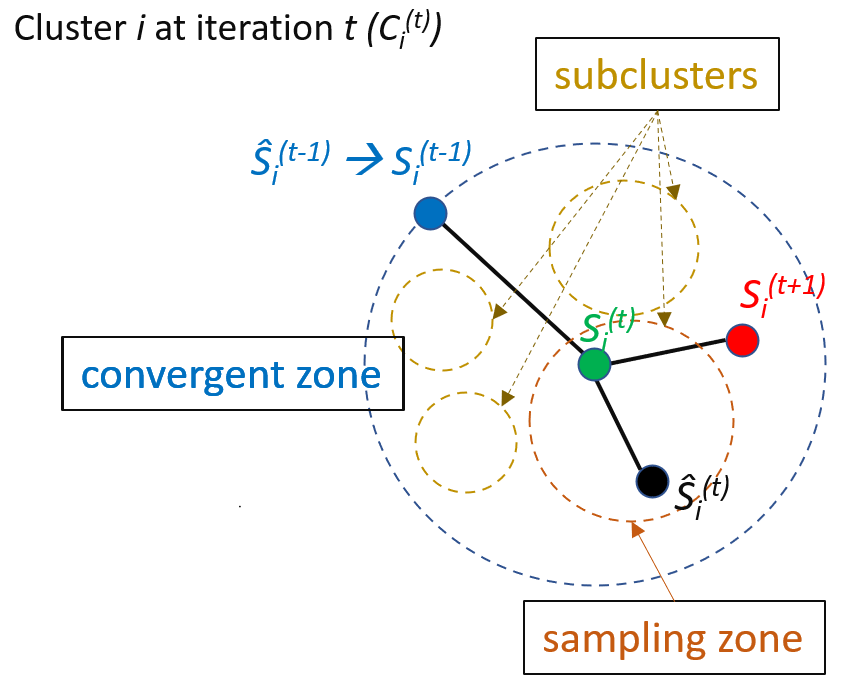}
\caption{Overview of SubCluster Guarantee Approach}
\label{fig:subcluster.jpg}
\end{figure}

We sample the centroid from the sampling zone using the ExpDP mechanism. Finally, we inject Laplace noise in the centroids of the clustering when our algorithm converges. It is because the differentially private centroids obtained are a subset of one of the local minima at which Lloyd's algorithm converges. The overview of the proposed approach can be seen in (\textbf{Figure : \ref{fig:subcluster.jpg}}). We show that a randomized iterative algorithm satisfies an invariant (given in the claim of Lemma 1) and always converges (\textbf{Proof: refer Lemma 1}). Finally, we show that the SubCluster algorithm is a randomized iterative algorithm that satisfies the invariant(given in \textbf{Lemma 1) (Proof: Refer Lemma 2}).

We have four main steps to update the centroids at each Lloyd step t.

\begin{enumerate}
    \item Let the differentially private centroid at iteration $t-1$ for a cluster $i$ be $\hat{S_i}^{(t-1)}$. Using this centroid, run one iteration of Lloyd's algorithm to get the current Lloyd's centroid ${S_i}^{(t)}$ for each cluster $i$.
    \item Using ${S_i}^{(t)}$ and ${S_i}^{(t-1)}$, generate a convergent zone for each cluster $i$ as described in \textbf{$Definition \:3$}.
    \item SubCluster the \textit{convergence zone} and sample one of the sub-cluster as our \textit{sampling zone} based on the probability assigned to each sub-cluster. The probability assignment is directly proportional to the number of points in each sub-cluster.
    \item Sample a differentially private $\hat{S_i}^{(t)}$ with EXpDP in the \textit{sampling zone} generated in step 3.
\end{enumerate}

\begin{figure*}[t]
\includegraphics[width=\linewidth]{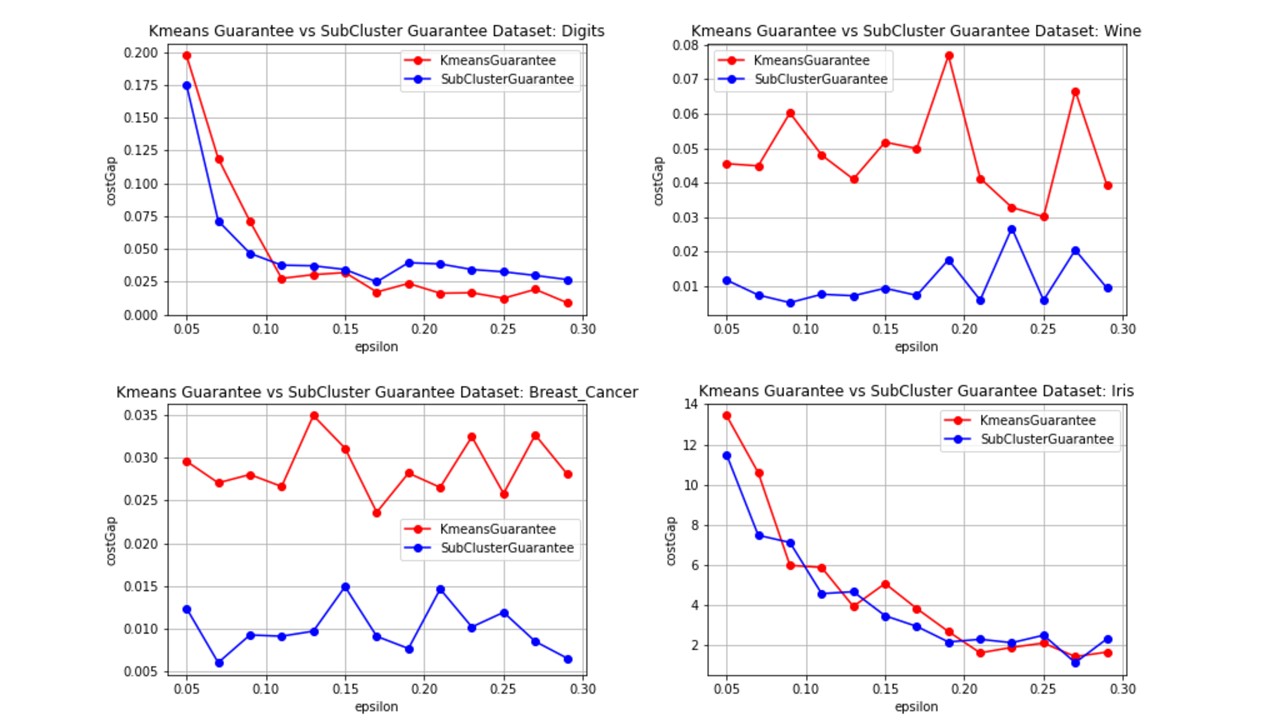}
\caption{Above figures plots the graph between costGap and epsilon budget for two approaches, the baseline as KmeansGuarantee and our approach SubClusterGuarantee. The algorithm was tested on four dataset, Digits (top-left), Wine (top-right), Breast Cancer (bottom-left), and Iris (bottom-right) datasets.}
\label{fig:DP_1.jpg}
\end{figure*}

Our approach surpasses the baseline approach in terms of clustering quality while maintaining the same DP requirements as that of the KMeans Guarantee approach, which is evident from the results obtained (\textbf{Figure : \ref{fig:DP_1.jpg}}). The better clustering quality is a result of our sub-clustering strategy to perturb centroid with a higher probability than the baseline approach towards the direction of the actual centroid generated by Lloyd's algorithm. The pseudo-code of our approach is shown in the \textbf{Algorithm 1 and Algorithm 2}.



\begin{figure*}[t]
\includegraphics[width=\linewidth]{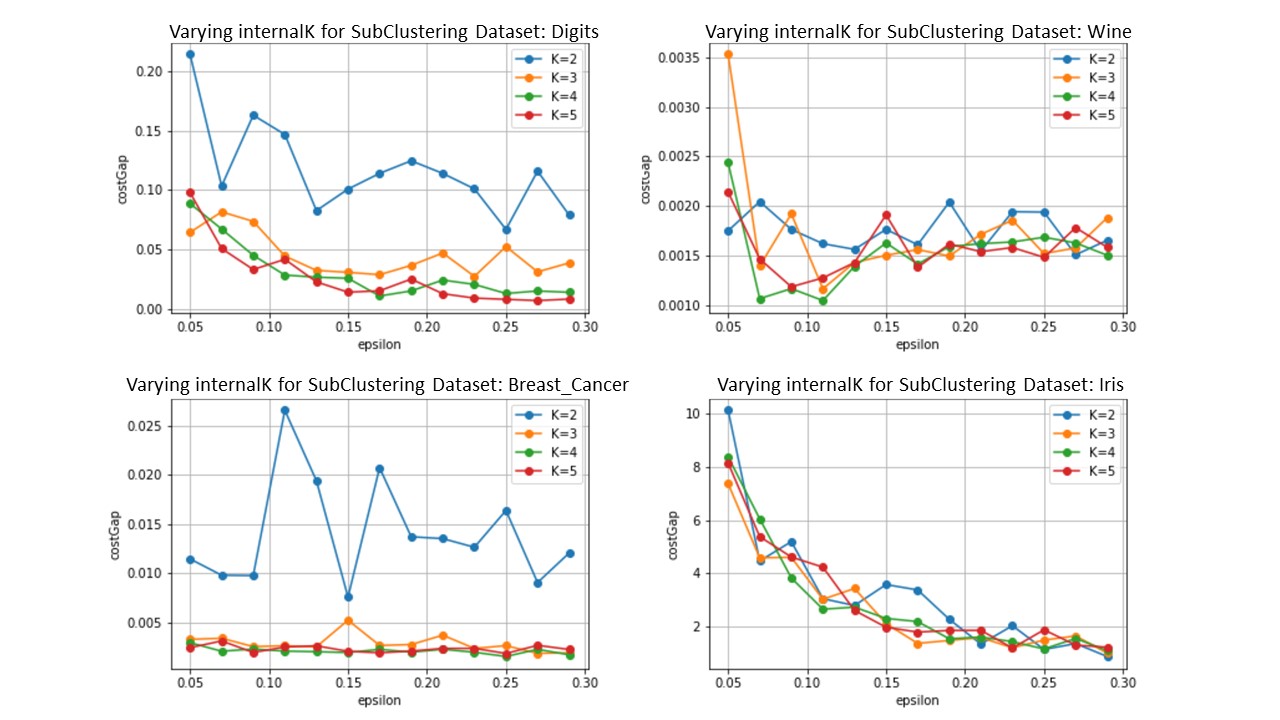}
\caption{Above figures plots the graph between costGap and epsilon budget for different internalK in SubClusterGuarantee Algorithm. The algorithm was tested for internalK=2,3,4,5 for all the four datasets, Digits (top-left), Wine (top-right), Breast Cancer (bottom-left), and Iris (bottom-right). Please note: K and internalK are the same parameter}
\label{fig:DP_2.jpg}
\end{figure*}

\textbf{Lemma 1: \cite{lu2020differentially}} A randomised iterative algorithm $\tau$ is convergent if, in ${{C}^{(t)}_{i}}$(Cluster i at iteration t), $\hat{S_i}^{(t)}$(sampled centroid using $\tau$), ${S_i}^{(t-1)}$(centroid before recentering) and ${S_i}^{(t)}$(centroid of ${{C}^{(t)}_{i}}$) satisfies the invariant, $\vert\vert \hat{S_i}^{(t)}-{S_i}^{(t)} \vert\vert < \vert\vert {S_i}^{(t)}-{S_i}^{(t-1)} \vert\vert$ in Euclidean distance, $\forall \space t$, $i$.

We reproduce this lemma from our baseline approach \cite{lu2020differentially}. Lemma1 and Lemma 2 together provides the completeness and proof for the convergence of our approach. If the distance between the sampled centroid ${\hat{S}^{(t)}_{i}}$ from the ${{C}^{(t)}_{i}}$ and the new centroid ${{S}^{(t)}_{i}}$ is less than the distance between the new ${{S}^{(t)}_{i}}$ and the old centroid ${{S}^{(t-1)}_{i}}$, then the random iterative algorithm will always converge. Intuitively, the loss of ${{C}^{(t)}_{i}}$ is minimum if the mean of ${{C}^{(t)}_{i}}$ is taken as centroid. But, if we slightly shift from the mean of ${{C}^{(t)}_{i}}$, then the loss will increase. However, if we can ensure that any sampled point from ${{C}^{(t)}_{i}}$ fulfills the condition: $\vert\vert\hat{S_i}^{(t)}-{S_i}^{(t)} \vert\vert < \vert\vert {S_i}^{(t)}-{S_i}^{(t-1)} \vert\vert$, it will lead to a lesser loss than ${{J^S}^{(t-1)}_{i}}$, thus, resulting into convergence of the randomised iterative algorithm. For the mathematical proof, refer \cite{lu2020differentially}.

\textbf{Lemma 2:} Differentially Private $k-$Means SubClustering approach (SubClustering) is a randomised iterative algorithm that satisfies the invariant $\vert\vert \hat{S_i}^{(t)}-{S_i}^{(t)} \vert\vert < \vert\vert {S_i}^{(t)}-{S_i}^{(t-1)} \vert\vert$.\\
\textbf{Proof:} SubClustering is an iterative algorithm that samples a set of centroids for each iteration with ExpDP mechanism, thus, making it a randomised iterative algorithm. It subclusters the points lying inside ${ConvergentZone}^{(t)}_{i}$. After subclustering, it samples one subcluster (sampling zone) with the assigned probabilities (linearly proportional to the number of data points in subcluster). Finally, it samples a datapoint from the sampled subcluster with ExpDP and call it as the centroid of ${ConvergentZone}^{(t)}_{i}$. Thus, our sampling zone always lies inside ${ConvergentZone}^{(t)}_{i}$. Therefore, the sampled point lies inside ${ConvergentZone}^{(t)}_{i}$ and it satisfies the invariant $\vert\vert \hat{S_i}^{(t)}-{S_i}^{(t)} \vert\vert < \vert\vert {S_i}^{(t)}-{S_i}^{(t-1)} \vert\vert$.

\section{Experimental Setup}
\subsection{Dataset Used}
We used following four datasets to test our work SubCluster Guarantee upon the baseline:
\begin{enumerate}
    \item \textbf{Iris \cite{asuncion2007uci}} dataset comprises total of 150 datapoints with four features and three classes. 
    \item \textbf{Wine\cite{asuncion2007uci}} dataset comprises total of 178 datapoints with 13 features and three classes. 
    \item \textbf{Breast Cancer\cite{asuncion2007uci}} dataset comprises total of 569 datapoints with 30 features and two classes. 
    \item \textbf{Digits\cite{asuncion2007uci}} dataset comprises of 1797 datapoints with 64 dimensions and 10 classes.
\end{enumerate}

\subsection{Metric for Clustering Quality}
To evaluate the clustering quality, we used the following equation to calculate the normalised difference between the differentially private algorithms (here, SubCluster Guarantee approach) $(Cost_{DP})$ and Lloyd's algorithm $(Cost_{Lloyd})$:
\begin{equation}
    Cost Gap = \frac{|Cost_{DP} - Cost_{Lloyd}|}{Cost_{Lloyd}} 
\end{equation} 
The smaller \textit{CostGap} \cite{lu2020differentially} represents the better quality of clustering. In the experiments, we compare the clustering quality of SubCluster Guarantee with KMeans Guarantee.

\section{Results and Discussion}
We tested our algorithm on four datasets. All the datasets have different dimensions ranging from 4 to 64 dimensions and training sets ranging from 150 to 1800. As defined in metric smaller gap represents the better clustering quality. From the (\textbf{Figure : \ref{fig:DP_1.jpg}}) we can observe that, cost gap for all the dataset is smaller or equal to the baseline. Thus, it is evident that our algorithm has better clustering quality than the existing work for all the datasets experimented. We varied internalK (parameter for number of sub-clusters) from 2 to 5.

Each experiment was conducted 30 times in the case of the Iris, Wine, and Breast cancer dataset and 10 times for digits dataset due to computational constraints. Finally, for each dataset, we took the average of all the experiments as our final result for plotting the graphs.

Comparing the SubCluster Guarantee (proposed approach) and K-means Guarantee approach (baseline) by taking an average of all the cost gaps for varied epsilon, and finally taking the ratio between K-means and SubCluster approach:
\begin{enumerate}
    \item In case of \textbf{Iris} dataset, the cost gap is 1.1 times smaller than baseline algorithm.
    \item In case of \textbf{Wine} dataset, the cost gap is 4.13 times smaller than baseline algorithm.
    \item In case of \textbf{Breast$\_$Cancer} dataset, the cost gap is 2.83 times smaller than baseline algorithm.
    \item In case of \textbf{Digits} dataset, the cost gap is almost same as that of baseline algorithm.
\end{enumerate}
\subsection{Detailed Analysis}
\begin{enumerate}
    \item \textbf{Iris}: Iris dataset has four dimensions and a very small training set of 150 data points. Our algorithm achieves better clustering quality than the baseline algorithm for smaller epsilon values. Since the number of data points is less in Iris, the impact of sub-clustering reduces, resulting in its performance similar to that of the baseline approach. From (\textbf{Figure : \ref{fig:DP_2.jpg}}), we can observe that changing the value of intenalK has a small impact on the costGap due to a small number of points in each sub-cluster. This is because there is a possibility that a sub-cluster has no data point when internalK is increased causing zero probability sub-cluster regions.
    \item \textbf{Wine}: The wine dataset has 13 dimensions and 178 data points in the training set. Our algorithm performs significantly better than the baseline, as observed in (\textbf{Figure : \ref{fig:DP_1.jpg}}). It is because the baseline algorithm is constrained to choose a theta in any abrupt direction ranging from $\left[-\pi/2,\pi/2\right]$ as shown in (\textbf{Figure : \ref{fig:kmeans.jpg}}). In contrast, our algorithm shifts the centroids in the direction where the future centroid of Lloyd's algorithm is more likely to move (in the expected case). From (\textbf{Figure : \ref{fig:DP_2.jpg}}), it is evident that internalK=4 for the wine dataset performs better than the rest of the internalK values. Here, the number of dimensions is more than Iris. Therefore, the spatial arrangement will be in an n-sphere which allows better sub-clustering.
    
    \item \textbf{Breast$\_$Cancer}: Breast$\_$Cancer dataset has 569 data points in its training set and 30 dimensions. Our algorithm performs exceptionally better than the baseline, with internalK equal to 4. From (\textbf{Figure : \ref{fig:DP_1.jpg}}), we can observe that there is no monotonous trend for the costGap. Trends are visible in other datasets due to the larger number of classification classes, whereas this dataset has only two classes. Thus, adding Laplace noise does not have a relation to the clustering quality. Increasing the internalK improves the clustering quality, with internalK being 4 having the least loss. It is because this dataset has a high number of dimensions and a larger number of training points than other datasets.

    \item \textbf{Digits}: It has 64 dimensions and 1797 data points in the training dataset. Although it has a large number of dimensions, our algorithm has a very small improvement over the baseline algorithm as seen in (\textbf{Figure : \ref{fig:DP_1.jpg}}). Because of the higher time complexity of our algorithm, it is hard to tune the internalK parameter. As the number of samples in a dataset increases, the internalK should increase because a single cluster can contain a large number of data points. But, due to limited computational resources, we were not able to experiment with it further. We took internalK to be 5 for our experiments as it performed best in the range $\left[2,5\right]$ as in the (\textbf{Figure : \ref{fig:DP_2.jpg}}). One of the intriguing findings in the dataset's results is that the curves based on the internalK have a clearly evident trend, which is a result of the large number of training data points. 
    
\end{enumerate}
Our proposed algorithm significantly improves over the baseline in terms of clustering quality, especially for the wine and breast cancer dataset. In addition our algorithm maintains the same DP requirements as that of existing works.

\section{Conclusion}

This work presents a novel method for improving the clustering quality of differentially private k-means algorithms while ensuring convergence. The novelty of our approach is the sub-clustering of the cluster to select the differentially private centroid, which has a higher probability of moving in the direction of the next centroid. 
We proved that our work surpasses the current state-of-the-art algorithms in terms of clustering quality. Especially for the Wine and Breast$\_$Cancer dataset, the clustering quality was significantly improved by 4.13 and 2.83 times than the baseline. In addition, we maintain the same DP requirements as that of baseline and other existing approaches.

\section{Future Work}
\begin{itemize}
\item In this work, we proved our claim using empirical results. We further plan to validate the results by providing mathematical bounds for the convergence degree and rate of the SubClustering Lloyd's algorithm. In terms of clustering quality, the proposed algorithm in this work is compared with k-means guarantee clustering only; to prove the effectiveness of our work, we plan to experiment with other algorithms in the literature including, PrivGene \cite{zhang2013privgene}, GUPT \cite{mohan2012gupt} and DWork \cite{dwork2011firm}.

\item The DP requirements in this work are the same as that of past literature, but in the future, we plan to explore ways to improve the current DP guarantees while maintaining the same clustering quality as in this work.

\item We used Exponential and Laplace mechanisms of DP in the proposed approach; we further plan to explore the third mechanisms, i.e., sample and aggregate framework, by integrating it with the current algorithm.

\item In our algorithm, the number of data points inside a cluster is variable. Thus we plan to choose an internalK, custom to the size of the cluster to improve the clustering quality.

\section*{Acknowledgement}
We would like to thank Prof. Anirban Dasgupta (IIT Gandhinagar) for his continuous support and guidance throughout the research.

\bibliography{main}

\end{itemize}





\end{document}